\documentclass[times, 10pt]{article}
\usepackage{times}
\usepackage{balance}
\usepackage{amsmath}
\usepackage{amssymb}
\usepackage{mathpazo}
\usepackage{hyperref}
\usepackage{multimedia}
\usepackage{amsfonts}
\usepackage{graphicx}

\newtheorem{theorem}{Theorem}

\newtheorem{lemma}[theorem]{Lemma}

\newtheorem{remark}[theorem]{Remark}

\newenvironment{proof}[1][Proof]{\noindent\textbf{#1.} }{\ \rule{0.5em}{0.5em}}
\begin{document}

\title{The steepest watershed : from graphs to images}

\author{Fernand Meyer \\
\emph{Centre de Morphologie Math\'{e}matique}\\
\emph{Mines-ParisTech, Paris, France}\\
\emph{fernand.meyer@mines-paristech.fr}\\
}

\maketitle
\thispagestyle{empty}

\begin{abstract}
The watershed transform is a powerful and popular tool for segmenting objects whose contours appear as crest lines on a gradient image : it associates to a topographic surface a partition into catchment basins, defined as attraction zones of a drop of water falling on the relief and following a line of steepest descent. To each regional minimum corresponds a catchment basin. Points from where several distinct minima may be reached are problematic as it is not clear to which catchment basin they should be assigned. Such points belong to watershed zones, which may be thick. Watershed zones are empty if for each point, there exists a unique steepest path towards a unique minimum. Unfortunately, the classical watershed algorithm accept too many steep trajectories, as they use too small neighborhoods for estimating their steepness. In order to nevertheless produce a unique partition they do arbitrary choices, out of control of the user. Finally, their shortsidedness results in unprecise localisation of the contours. We propose an algorithm without myopia, which considers the total length of a trajectory for estimating its steepness ; more precisely, a lexicographic order relation of infinite depth is defined for comparing non ascending paths and chosing the steepest. For the sake of generality, we consider topographic surfaces defined on node weighted graphs. This allows to easily adapt the algorithms to images defined on any type of grids in any number of dimensions. The graphs are pruned in order to eliminate all downwards trajectories which are not the steepest. An iterative algorithm with simple neighborhood operations performs the pruning and constructs the catchment basins. The algorithm is then adapted to gray tone images. The neighborhood relations of each pixel are determined by the grid structure and are fixed ; the directions of the lowest neighbors of each pixel are encoded as a binary number. Like that, the graph may be recorded as an image. A pair of adaptative erosions and dilations prune the graph and extend the catchment basins. As a result one obtains a precise detection of the catchment basin and a graph of the steepest trajectories. A last iterative algorithm allows to follow these downwards trajectories in order to detect particular structures such as rivers or thalweg lines of the topographic surface.

\end{abstract}

\section{Introduction}

The watershed line or divide line of a topographic surface is the boundary
separating its catchment basins.\ A drop of water falling on this surface
glides along a line of steepest descent until it is captured by a regional
minimum. A catchment basin is the attraction zone of a minimum.\ Catchment
basins may overlap and the overlapping zone is precisely a divide line, as a
drop of water falling on it may glide towards several distinct minima.\ Any
gray tone image may be considered as a topographic surface, where the altitude
is proportional to the gray-tone.\ Consider in particular the gradient of an
image to segment.\ In absence of noise or texture, the inside of the objects
and of the background appears as minima and the contours appear as crest lines
separating the minima. Each object, appearing in the gradient image as the
catchment basin of a minimum, is easily extracted by the watershed transform.
Each minimum gives birth to a catchment basin ; with two many meaningless
minima, over-segmentation occurs.\ Marker based segmentation regularizes the
gradient image ; it consists in flooding the topographic surface in order to
keep only one regional minimum for each object of interest \cite{beucher79}.
More generally, closing reconstructions or floodings regularize the gradient
as they completely fill some catchment basins up to their lowest pass point.
As a result these catchment basins are absorbed by neighboring basins,
yielding a coarser segmentation. To a series of increasing floodings will be
associated a hierarchy \cite{floodingsegmey}. The filling of lakes and
subsequent absorption of their catchment basins by neighboring regions may be
ordered according some geometric criteria, such as the depth of the lakes
\cite{grimaud92}, \cite{Najman94}, \cite{saliency}, their area or their volume
\cite{vachier95},\cite{vachier95b}.\ They may also be filled in an interactive
mode and be a building block for interactive segmentation \cite{xisca2002}.

The watershed being a powerful tool for segmentation has been victim of its
success : a number of definitions and algorithms have been published, claiming
to construct a watershed line or catchment basins, although they obviously are
not equivalent. It is out of the scope of the present paper to give an
exhaustive bibliography of the concept of watershed. Distinct algorithms
published in the literature will produce different results.\ But even the same
algorithm may produce distinct results if one changes the scanning order of
the image.\ Jos.\ Roerdink gives a review of the most popular watershed
definitions and implementations \cite{Roerdink01thewatershed}. An historical
analysis with many references on how the watershed idea has been developed,
triggered both by theoretical considerations and by technological
possibilities for its implementation may be found in \cite{wshedhistory}.

We propose an algorithm for the following situation : a gray tone image is
given, represented on a grid or on a graph and we desire delineating its
catchment basin. Moreover we desire constructing a partition into catchment
basins, often at the price of arbitrary divisions of the zones where they overlap.
We will compare the performances between algorithms dealing with the same
situation.\ These algorithms do not cover the situation where a topographic surface
is constructed by a shortest path algorithms and represents for each pixel its
distance to predefined seeds as in \cite{meyer92}, \cite{IftLotufo}. The
Vorono\"{\i} tessellation associated to this distance also constitutes a
partition ; furthermore, if one applies the shortest path algorithm by
Dijkstra, one may assign to each node one ancestor, through which its distance
to the nearest neighbor has been constructed. The algorithm produces a minimum
spanning forest.

Further we do not consider the algorithms which construct a thin
watershed line separating catchment basins.\ In a hierarchy one goes from a
fine to a coarse partition by merging adjacent regions. This operation is
immediate if one deals with partitions : one assigns to all regions to be
merged the same label.\ It is however more problematic if the contour is
materialized between the regions and paradoxical situations may be met if one
does not carefully chose the graph representing the images \cite{fusiongraphs}%
. This is an additional reason why to prefer watershed zones without
boundaries between regions ; furthermore representing contours wastes space in
the image and makes it impossible to segment adjacent small structures.

As a conclusion, we consider here the watershed transforms associating to a
topographic surface a partition into catchment basins, defined either by water
streaming down or flowing up. The first definition defines the catchment
basins as the domain of attraction of a drop of water falling on the surface
and gliding downwards to reach a minimum. The second floods the domain from
sources placed at the minima ; lakes containing distinct minima meet along the
divide line. The progression of the rain or of the flood follows lines of
steepest descent. Whether a drop of water glides down a surface or a flood
progresses upwards finally has no importance, the result will be the same if
they follow identical trajectories.\ Among all non increasing trajectories
between a node and a minimum, the algorithm has to chose the steepest ; if
several equivalent trajectories connect a node with distinct minima, this node
belongs to two overlapping catchment basins ; if the target is a partition,
we have to divide the overlapping zone between the two adjacent basins. It is
not clear and often out of control of the user, how this division is made by
the algorithms published in the literature. The resulting partition may
dramatically vary if one applies one type of division rules or another.\ The
same algorithm may also give different results if one simply changes the scanning
order of the image.\ This makes the positioning of the contours is neither
stable nor precise.\ 

Ideally, if each pixel had only one downwards trajectory towards a unique
minimum, catchment basins would not overlap.\ In the present work we do not
reach complete unicity but get as close as possible to this ideal situation.
We propose a method which selects the steepest trajectories possible. We
propose an algorithm without myopia, which considers the total length of a
trajectory for estimating its steepness. It turns out that for natural images
these trajectories often are unique.\ 

For the sake of generality we develop the algorithm for node weighted
graphs.\ The result may then easily be transposed to images defined on
arbitrary grids and any number of dimensions.

The outline of the paper is the following. We first consider topographic
surfaces defined on node weighted graphs.\ The graphs are pruned in order to
eliminate all downwards trajectories which are not the steepest.\ An iterative
algorithm with simple neighborhood operations performs the pruning and
constructs the catchment basins.\ The algorithm is then adapted to gray tone
images. The graph structure itself is encoded as an image thanks to the fixed
neighborhood structure of grids.\ A pair of adaptive erosions and dilations
prune the graph and extend the catchment basins. As a result one obtains a
precise detection of the catchment basins and a graph of the steepest
trajectories. A last iterative algorithm allows to follow selected downwards
trajectories in order to detect particular structures such as rivers or
thalweg lines of the topographic surface.

\section{The watershed on weighted graphs}

\subsection{Weighted graphs}

A \textit{non oriented graph} $G=\left[  N,E\right]  $ is a collection of
vertices or nodes $N$ and of edges $E$ ; an edge $u\in E$ being a pair of
vertices \cite{berge85},\cite{gondranminoux}. Each node $\nu_{i}$ is
weighted with a real number $w(\nu_{i})$.\ 

The \textit{subgraph} spanning a subset $A\subset N$ is defined as
$G_{A}=[A,E_{A}]$, where $E_{A}$ are the edges linking two nodes of $A.$

The \textit{partial graph} associated to a subset of edges $E^{\prime}\subset
E$ has the same nodes as $G$ but has less edges ; it is defined by $G^{\prime
}=[N,E^{\prime}].$

A gray tone image may be considered as a graph, in which each pixel $i$
becomes a node, with a weight $w(i)$ representing its gray tone. Neighboring
pixels are linked by an edge.\ 

A subgraph $G^{\prime}$ of a node weighted graph $G$ is a \textit{flat zone},
if any two nodes of $G^{\prime}$ are connected by a path where all nodes have
the same altitude.

A subgraph $G^{\prime}$ of a graph $G$ is a \textit{regional minimum} if
$G^{\prime}$ is a flat zone and all neighboring nodes have a higher altitude.

\subsection{Drainage graphs and their catchment basins}

We associate to $G$ an oriented graph $\overrightarrow{G}$, called drainage
graph, encoding all possible directions a drop of water may follow when
falling on a given node $i$ outside a regional minimum.\ If $j$ is one of the
lowest neighboring nodes of $i$ in $G$, an arc is created in $\overrightarrow
{G}$ from $i$ towards $j.\ $If $i$ and $j$ both do not belong to regional
minima and have the same weights, then two arcs are created, one from $i$ to
$j$ and one from $j$ to $i.\ $

A \textit{drainage path }between two nodes $x$ and $y$ is a sequence of nodes
$(\nu_{1}=x,\nu_{2},...,\nu_{k}=y)$ such that two successive nodes $\nu_{i}$
and $\nu_{i+1}$ are linked by an arc.\ 

\begin{lemma}
From each node outside a regional minimum starts a drainage path to a regional minimum
\end{lemma}

\begin{proof}
Any node $i$ outside a regional minimum has a lower neighboring node, if it
does not belong to a plateau. Otherwise it belongs to a plateau, containing
somewhere a node $j$ with a lower neighboring node, as the plateau is
not a regional minimum.\ The plateau being connected, there exists a path of
constant altitude in the plateau between $i$ and $j$.\ Following this path, it
is possible, starting at node $i$ to reach the lower node $k.$ This shows that
for each node there exists an oriented path to a lower neighboring node.
Taking this new node as starting node, a still lower node may be reached. The
process may be repeated until a regional minimum is reached.
\end{proof}

Thanks to this lemma, it is possible to define the catchment basins of the minima.\ 

The \textbf{catchment basin} of a minimum $m$ is the set of all nodes from
which starts an oriented path towards $m$ in $\overrightarrow{G}.$

\textbf{The watershed zones }are the nodes belonging to more than one
catchment basin. 

If a node has several lowest neighbors, each arc towards one of them is the
first arc of a drainage path. Each lowest neighbor in turn may have several
second lowest neighbors, multiplying the number of drainage paths with the
same origin ; if two such paths link a node $i$ with two distinct minima, then
$i$ belongs to a watershed zone.\ 

\textbf{The restricted catchment basins} are the nodes which belong to only
one catchment basin : it is the difference between the catchment basin of a
minimum $m$ and the union of catchment basins of all other minima. From a node
in a restricted catchment basin, there exists a unique non ascending path
towards a unique regional minimum.\

\subsection{Steepest drainage paths}

For some nodes in a drainage graph $\overrightarrow{G},$ there exist several
oriented paths towards regional minima. If a node has several lower neighbors,
each arc towards one of them may be the first arc of such a path. Each lowest
neighbor in turn may have several lowest neighbors, multiplying the choices of
drainage paths.\ This is the reason why thick watershed zones appear, as
several oriented paths, starting from the same origin node may reach distinct
regional minima.\ 

The number of drainage paths with the same origin is reduced if we consider
not only the first neighbors of each node, but larger neighborhoods. Suppose
that a node $i$ has two lowest neighbors $j$ and $k$ ; each arc $(i,j)$ and
$(i,k)$ is the first arc of an oriented path towards a minimum. Suppose now
that the lowest neighbor of $j$ is lower than the lowest neighbor of $k,$
indicating that the path passing through $(i,j)$ is steeper than the path passing
through $(i,k)$ : both have identical weights on the first two nodes, but the
third node differenciates them.\ More generally, two paths may have identical
weights on the $k$ first nodes and distinct nodes on the node $k+1.$ A lexicographic
order relation will be defined for comparing the steepness of non ascending paths.
steepness of a non ascending path will be based on the lexicographic order.\ The weights of
the nodes along an oriented drainage path, starting from a given node and
following the path downwards, are by construction a series of never increasing
weights.\ A lexicographic order may be defined for comparing paths: a path
$(\nu_{1},\nu_{2},...\nu_{k})$ is steeper than a path $(\nu_{1}^{\prime}%
,\nu_{2}^{\prime},...\nu_{l}^{\prime})$ if $w(\nu_{1})<w(\nu_{1}^{\prime})$ or
if there exists an index $t$ such that for $i<t,$ we have $w(\nu_{i}%
)=w(\nu_{i}^{\prime})$ and $w(\nu_{t})<w(\nu_{t}^{\prime}).$

The number of distinct paths with the same origin towards distinct minima
sharing exactly the same list of non increasing weights is extremely
low and often reduced to one if one considers natural images.\ This shows that
if one considers only the paths with maximal steepness in a drainage graph,
the size of the watershed zones is strongly reduced. They will be empty, if there
exists only one path with maximal steepness linking each node with a regional minimum.

\subsection{Pruning drainage graphs}

If a path $\pi=(\nu_{1},\nu_{2},...\nu_{k})$ of a drainage graph has a maximal
steepness, then any sub-path $(\nu_{l+1},\nu_{l+2},...\nu_{k})$ obtained by
skipping the first $l$ nodes is still a path of maximal steepness.\ If another
path $\pi^{\prime}$ of origin $\nu_{l+1}$ were steeper, then concatenating
$(\nu_{1},\nu_{2},...\nu_{l})$ and $\pi^{\prime}$ would produce a path steeper
than $\pi.$ This lemma has a consequence: if $(i,j)$ is not the first arc of a
steepest path with origin $i,$ then no other steepest path will pass through
this edge. As a matter of fact, if $\pi$ were a steepest pass passing through
$(i,j),$ the sub-path of $\pi$ starting  at $i$ would be a steepest path, which is not
the case.

The previous analysis shows that if we cut all arcs of a drainage graph which
are not the highest edge of a steepest path, we obtain a partial graph with
exactly the same steepest paths as the original one. For cutting the highest
edge of each non steepest path, the naive approach would compare the paths
with the same origin two by two, follow each one downwards until two edges are
found with distinct weights : the path leading to the highest edge would be
the pass with a lower steepness and its initial edge could be pruned.

A more clever pruning algorithm relies on the erosion $\varepsilon$ on the
drainage graph, which lets the isolated regional minima nodes unchanged and
assigns to each other node the weight of its lowest neighbors to which it is
linked by an arc. Consider two paths $\pi=(\nu_{1},\nu_{2},...\nu_{k})$ and
$\pi^{\prime}=(\nu_{1},\nu_{2}^{\prime},...\nu_{k}^{\prime})$ with identical
weights for the $k-1$ first nodes and verifying $\nu_{k}>\nu_{k}^{\prime},$
indicating that $\pi^{\prime}$ is steeper than $\pi.$ Eroding the graph
$\overrightarrow{G}$ once assigns to the node $\nu_{1}$ the weight $w(\nu
_{2})=w(\nu_{2}^{\prime})$ and to the nodes $\nu_{2}$ and $\nu_{2}^{\prime}$
the weights $w(\nu_{3})=w(\nu_{3}^{\prime}).\ $With successive erosions, the
weights of the nodes along the paths $\pi$ and $\pi^{\prime}$ file past the
two first nodes.\ The erosion $k-1$ assigns to the node $\nu_{1}$ the weight
$w(\nu_{k-1})=w(\nu_{k-1}^{\prime})$ and to the nodes $\nu_{2}$ and $\nu
_{2}^{\prime}$ respectively the weights $w(\nu_{k})$ and $w(\nu_{k}^{\prime})$
which are different.\ Hence, after $k-1$ erosions, the graph $\varepsilon
^{(k-1)}\overrightarrow{G}$ is not a drainage graph anymore, as the edge
$(\nu_{1},\nu_{2})$ links $\nu_{1}$ with $\nu_{2}$ which is not one of its
lowest neighbor.\ This edge should thus be pruned in order to obtain a partial
graph which still is a drainage graph. This cutting achieves the intended
pruning : the first edge of a non steepest path is cut.\ 

\begin{figure}
[ptb]
\begin{center}
\includegraphics[width=0.90\textwidth]
{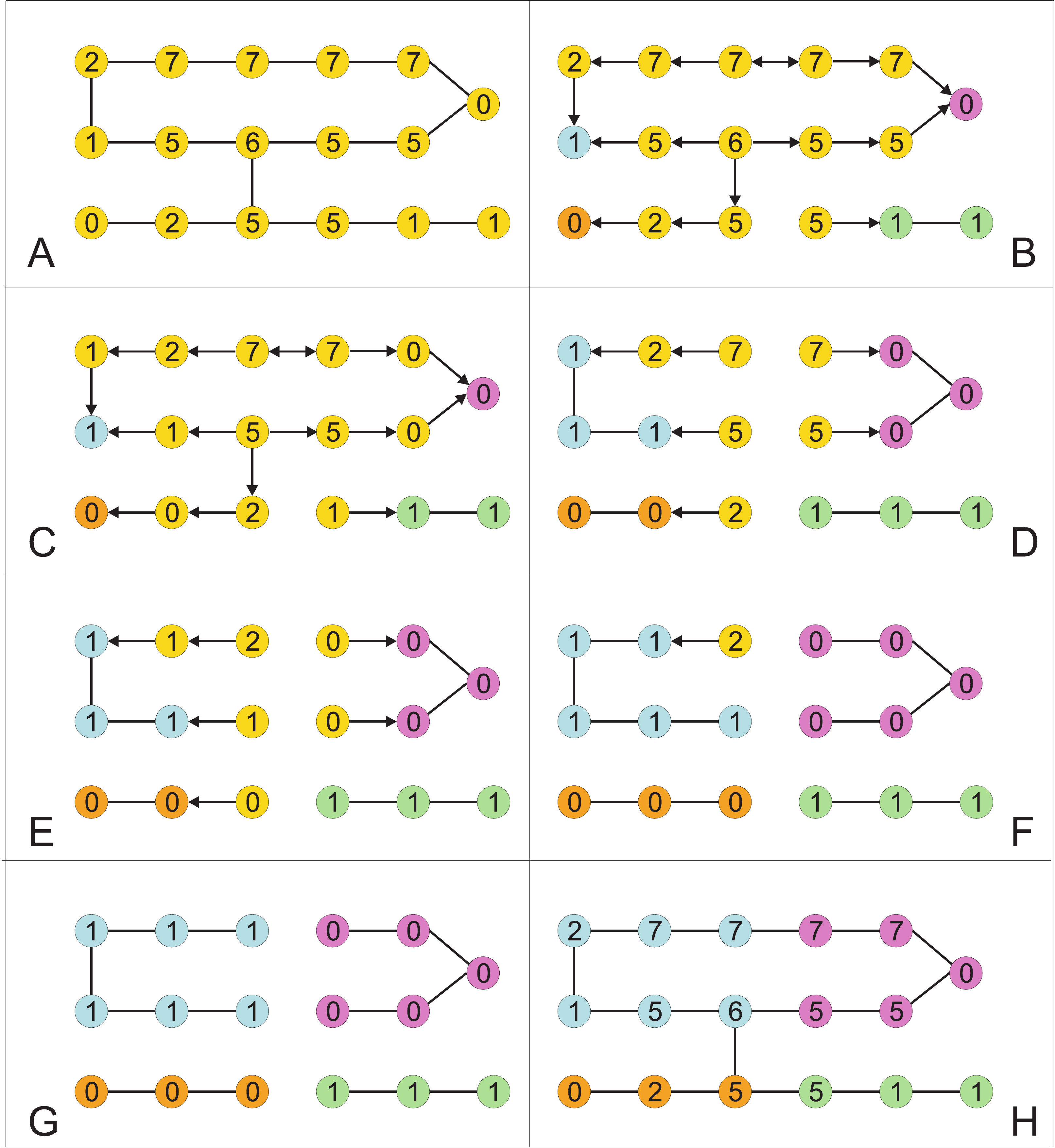}%
\caption{Construction of the catchment basins of a node weighted graph}%
\label{drainage2}%
\end{center}
\end{figure}

This leads to the following algorithm for obtaining the steepest drainage
graph, starting with an edge weighted graph $G$:\newline* construct a drainage
graph $\overrightarrow{G}$ by linking by an arc each node of $G$  outside the regional minima with each of
its lowest neighbors. Detect, label the regional minima and replace their
inside arcs by edges.\ Figure \ref{drainage2}\_B represents the drainage graph
of fig.\ref{drainage2}\_A.\newline* Repeat until stability : a) erode the
graph $\overrightarrow{G}$ and cut all arcs linking a node to another which is
not one of its lowest neighbors ; b) if $(i,j)$ is an oriented arc from $i$ to
$j,$ and if $j$ holds a label whereas $i$ has no label, then $i$ is assigned
the label of $j$, the arc between $i$ and $j$ is replaced by an edge and all
arcs with origin $i$ are suppressed. If $i$ points to several nodes with
distinct labels, one of them is chosen and the same procedure applied.\\ 
Applied on fig.\ref{drainage2}\_B the erosion and pruning produces
fig.\ref{drainage2}\_C ; the label propagation produces fig.\ref{drainage2}%
\_D\ .\ The next iteration produces fig.\ \ref{drainage2}\_E and
\ref{drainage2}\_F.\ After the last erosion, pruning and final label propagation, all
nodes are labeled and the steepest drainage graph obtained in fig.\ \ref{drainage2}\_G.\ The final
partition is superimposed on the initial graph in fig.\ref{drainage2}\_H.

\section{The watershed on images}

\subsection{Representing an oriented graph for images.}

In order to transpose to images the preceding algorithm initially defined for graphs,
one has to find a representation of the drainage graph itself.\ The nodes are
simply the pixels of the image to which the watershed algorithm is
applied.\ The nodes hold three types of valuations and will be represented on three images. 
First, a gray tone image holding the initial distribution
of gray tones and its evolution as the algorithm proceeds. The second image holds the labels of the minima and
of the catchment basins as they expand. The
last image is more original as it has to encode the drainage graph itself.

\begin{figure}
[ptb]
\begin{center}
\includegraphics[width=0.45\textwidth]
{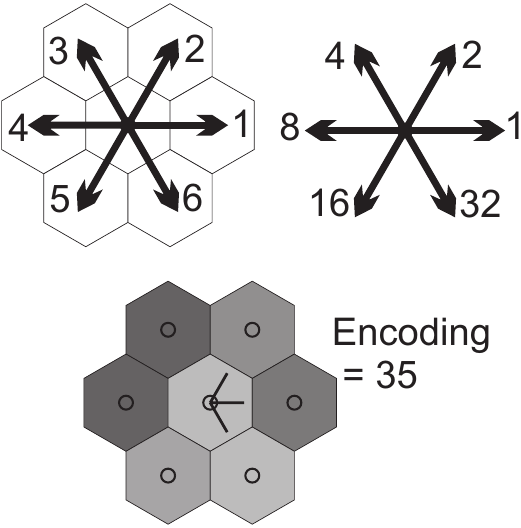}%
\caption{The encoding of the directions of the neighbors in the hexagonal
raster, and the weights of the corresponding bits in the binary representation
of the arrows. An example with three arrows with weights 2, 1 and 32 is
represented by the binary number 100011, i.e.\ the decimal number 35.}%
\label{wfal92}%
\end{center}
\end{figure}

The grids on which images are represented have a regular structure, where each
node has the same number of neighbors, in identical positions.\ Numbering the
neighbors according to their direction allows to represent the neighborhood
relation of each pixel with a binary number, where each bit encodes for one
direction.\ The n-th bit is set to 1 if and only if there exists an arrow
between the central point and its n-th neighbor ; such oriented arrows having
as origin a node $i$ are called \textit{out-arrows} of $i$. Figure
\ref{wfal92} shows the numbering of the directions in a hexagonal raster and
the corresponding bit planes (on the right).\ The bottom image gives an
example of encoding a particular neighborhood configuration : 2 +\ 1\ +\ 32\ =
35.\ This representation has been introduced by F.\ Maisonneuve in his seminal
work on watersheds \cite{FrancisM}.\

\begin{figure}
[ptb]
\begin{center}
\includegraphics[width=0.6\textwidth]
{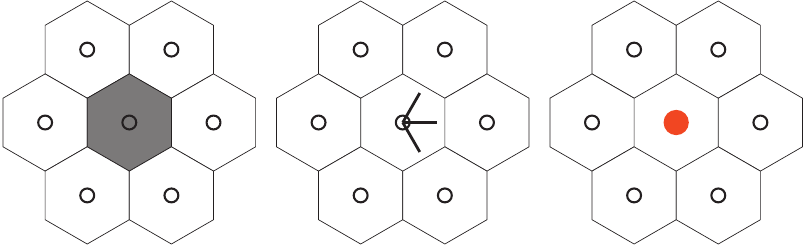}%
\caption{The three images used for constructing the catchment basins : a gray
tone image, an image of arrows and an image of labels.\ }%
\label{wfal94}%
\end{center}
\end{figure}

The algorithm creates and updates 3 images: 1) the gray tone image itself, 2)
an image of labels representing the regional minima and the catchment basins
in construction, 3) the encoding of the \textit{out-arrows} representing the
arcs of the drainage graph. \  Fig. \ref{wfal94} presents the three images. Fig. \ref{wfal94}.1 represents the gray tone
image. Fig. \ref{wfal94}.2 represents the image of out-arrows encoding the drainage graph. 
Fig. \ref{wfal94}.3 represents a label at the central position, represented as a colored dot.

\begin{figure}
[ptb]
\begin{center}
\includegraphics[width=0.8\textwidth]
{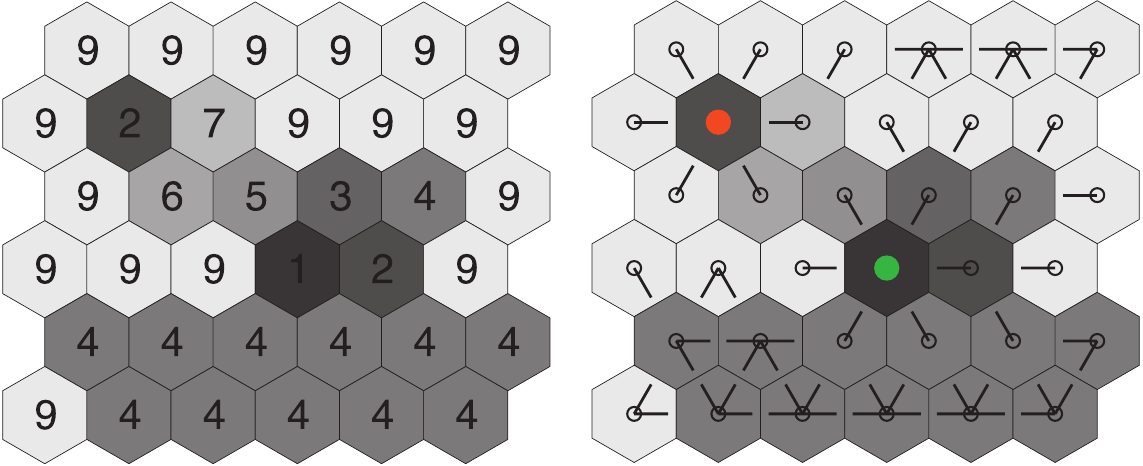}%
\caption{A gray tone image, the arrows representing its drainage graph and the
labels of the regional minima.\ }%
\label{wfal96}%
\end{center}
\end{figure}

\ Fig. \ref{wfal96}.1 presents a gray tone
image.\ Fig. \ref{wfal96}.2 combines the three images: the gray tone value for
each node, a colored dot representing the label of the minima and the initial
\textit{out-arrows} encoding the arcs.

\subsection{An adaptive erosion and dilation, guided by the arrows}

The successive prunings of arrows are easily implemented with two neighborhood
transformations, the first being an adaptive erosion for propagating the gray
tone values and pruning the arrows, the second being an adaptive dilation for
propagating the labels of the regional minima as they are extended.\ 

\subsubsection{An adaptive erosion}

\begin{figure}
[ptb]
\begin{center}
\includegraphics[width=0.4\textwidth]
{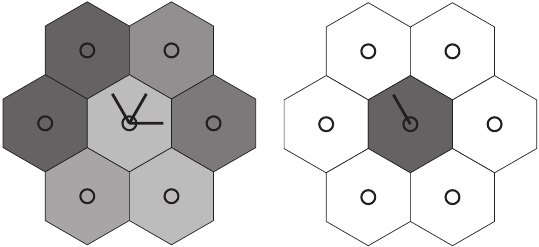}%
\caption{An example of adaptive erosion and pruning.\ }%
\label{wfal80}%
\end{center}
\end{figure}

We define an adaptive erosion and combined pruning on the arrowed image. It uses 
and updates both the gray tone image as the arrows image. If a
pixel is without arrows, it is left unchanged.\ Otherwise, it is replaced by
the lowest of its arrowed neighbors and only the arrows towards these
neighbors are kept.\ In figure \ref{wfal80} a pixel has two lowest neighbors,
but only one of them is arrowed.\ Hence only the arrow towards this node is kept.%

\begin{figure}
[ptb]
\begin{center}
\includegraphics[width=0.4\textwidth]
{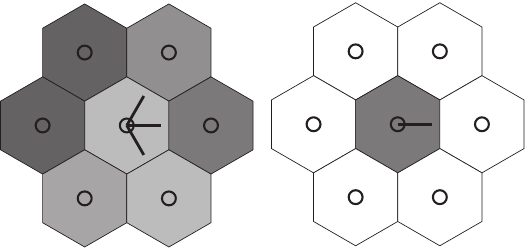}%
\caption{Adaptive erosion and pruning}%
\label{wfal85}%
\end{center}
\end{figure}

In figure \ref{wfal85} the two lowest neighbors are not arrowed : they are
discarded and only the lowest arrowed neighbor is taken into consideration :
its value is assigned to the central pixel and only the arrow towards it is kept.\ %

\begin{figure}
[ptb]
\begin{center}
\includegraphics[width=0.4\textwidth]
{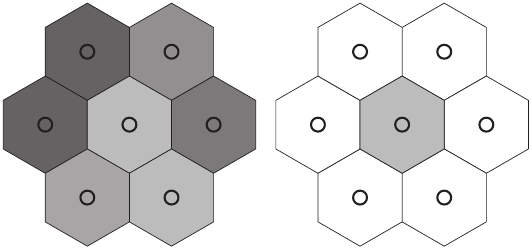}%
\caption{In the absence of any arrow, the central pixel is left unchanged.}%
\label{wfal86}%
\end{center}
\end{figure}

When no arrow is present, the central pixel is left unchanged as in figure
\ref{wfal86}.\ %

\subsection{An adaptive dilation}

The adaptive dilation propagates the labels of the regional minima. It
modifies both the image of labels and of arrows. It is guided by the
drainage graph. 

Recall that the labels are represented by strictly positive values, the pixels
without labels having the value $0$ in the labeled image. Furthermore, the
algorithm cares for the fact that labeled pixels have no arrows.\ This is true at
initialization, when the regional minima get their labels. It is also true
during the expansion of the catchment basins, as a pixel loses its arrows as
soon it gets a label.\ 

The label propagation is done by an adaptive dilation of the label images
guided by the arrows image. One considers the pixels without labels (a pixel
with a label has no out-arrows) ; such a pixel gets the highest label of its
arrowed neighbors.\ In the absence of arrowed and labeled neighbors, this
value is 0. The operation is thus an adaptive dilation.\ Furthermore, every time a pixel gets a label, its arrows are
suppressed, as it now belongs to a catchment basin.\ This produces an
additional pruning of the drainage graph.\ Below we illustrate the combination
of the adaptive erosion and dilation in a number of situations

\begin{remark}
1) Neighboring nodes of the same catchment basins are not linked by an edge ;
they are identified by the label they hold.
\newline2) In case where a pixel
without label has 2 or more out-arrows towards distinct labeled pixels, 
only one of them has to be chosen in order to get a partition.\ 
Chosing the highest of them constitutes an adaptive dilation. .
 As a matter of fact, we may chose arbitrarily one of them.
This is the only place where a choice takes place in the algorithm.\ It
divides the catchment zones and produces a partition. Such situations are
nevertheless rare, as we propagate the labels along the trajectories whose 
steepness is estimated by taking into account their total length. The necessity of 
a choice appears only in the case where two trajectories have exactly the same distribution of
node weights, from top to bottom.
\end{remark}

\subsection{Combination of the adaptive erosion and dilation : illustration}

\begin{figure}
[ptb]
\begin{center}
\includegraphics[width=0.6\textwidth]
{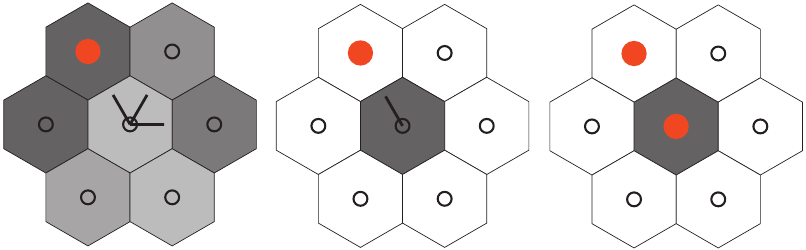}%
\caption{Adaptative erosion, pruning and guided dilation of the labels\ }%
\label{wfal83}%
\end{center}
\end{figure}

\begin{figure}
[ptb]
\begin{center}
\includegraphics[width=0.6\textwidth]
{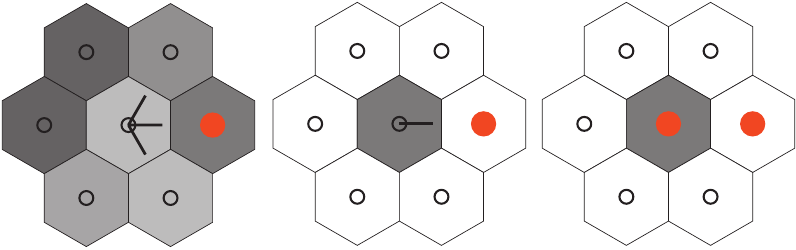}%
\caption{Adaptative erosion, pruning and guided dilation of the labels\ }%
\label{wfal81}%
\end{center}
\end{figure}

\begin{figure}
[ptb]
\begin{center}
\includegraphics[width=0.6\textwidth]
{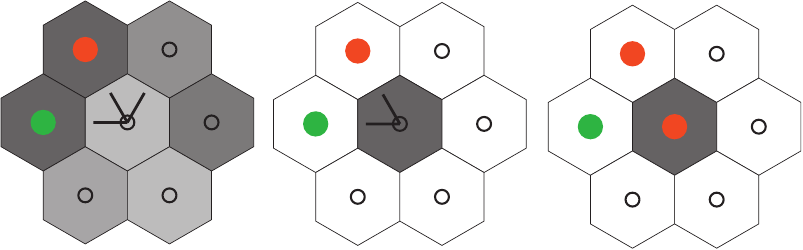}%
\caption{In the case where two or more distinct labels are present in the direction of
arrows, the highest of them, or one, chosen arbitrarily, is propagated}%
\label{wfal84}%
\end{center}
\end{figure}

\begin{figure}
[ptb]
\begin{center}
\includegraphics[width=0.4\textwidth]
{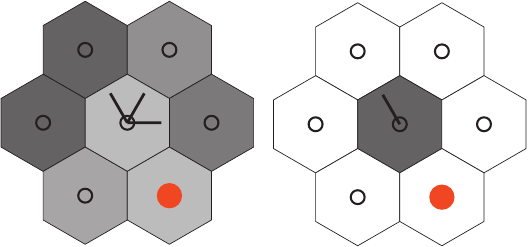}%
\caption{Case where a labeled pixel is present in the neighborhood of the
central pixel, but as it is not arrowed, it is not propagated to the central
pixel.}%
\label{wfal87}%
\end{center}
\end{figure}

The following figure shows how the adaptive erosion and dilation are used in
sequence. The erosion assigns to the central pixel the value of its lowest
arrowed neighbor ; only the arrow towards this neighbor is kept. And when this
arrow points towards a labeled neighbor, this label is propagated to the
central pixel by the adaptive dilation, and its arrows are
suppressed.\ Figures \ref{wfal83},\ref{wfal81},\ref{wfal84},\ref{wfal87}
present how, in different neighborhood configurations, the combined adaptative
erosion and dilation erode the gray tone, prune the arrows and propate the labels.\ %

%

%

%

\subsection{The complete watershed algorithm}

\begin{figure}
[ptb]
\begin{center}
\includegraphics[width=0.99\textwidth]
{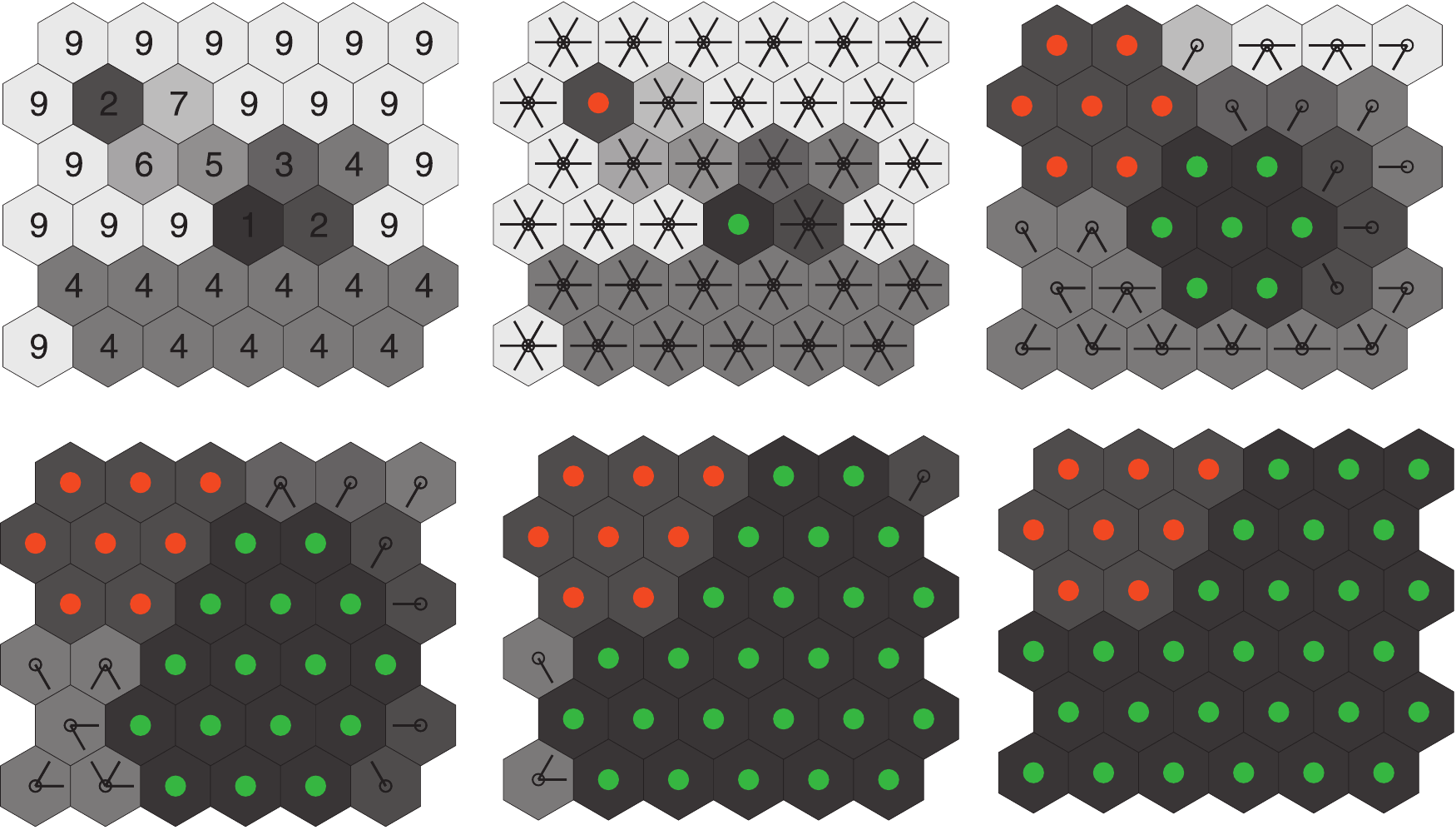}%
\caption{Initial gray tone distribution, labeling of the regional minima and
arrowing in all directions, followed by 4 iterations of a combined adaptive
erosion and dilation. }%
\label{wfal67}%
\end{center}
\end{figure}

Figure \ref{wfal67} shows that the algorithm can be initialized with arrows in
all directions. After the first combined adaptive erosion and dilation, only the
arrows towards the lowest neighbors are kept and the labels propagated
accordingly. After the initialization phase, the combined erosion of gray
tones, pruning of arrows and dilation of labels is iteratively applied until
the labels cover the total domain.\ Convergence is attained after 4 iterations
for figure \ref{wfal67}.%

\begin{figure}
[ptb]
\begin{center}
\includegraphics[width=0.99\textwidth]
{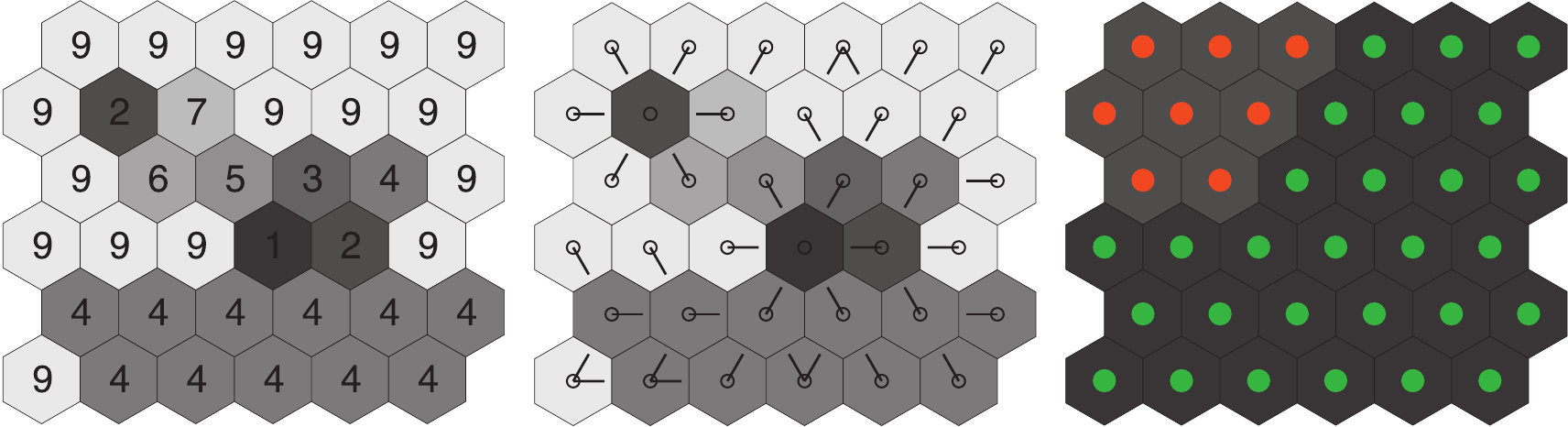}%
\caption{Final catchment basins, and recording of the last arrow distribution
of each node before it gets its label.\ }%
\label{wfal88}%
\end{center}
\end{figure}

By recording the arrows existing for each pixel, just before it gets its
label, one obtains the final and steepest drainage graph, where all prunings have been done as
illustrated in fig. \ref{wfal88}.\ The trajectories of a drop of water falling
on the surface are extremely selective and narrow.\ We will use this selectivity in the
last part of the paper for following lines of steepest descent and thalweg lines on a topographic surface.\ %

\subsection{Successive steps of the pruning}

During the successive adaptive erosions, no choice is ever made.\ 
A choice may appear necessary when two  when two distinct
arrowed and labeled pixels are present in the neighborhood of a pixel. 
The adaptive dilation choses the highest of them. Other rules may be introduced, 
as for instance a random choice. The occurence of such choices is rare in natural images, as they 
only happen when two distinct drainage paths linking a node with two distinct minima has
exactly the same distribution of weights.

%

\begin{figure}
[ptb]
\begin{center}
\includegraphics[width=0.66\textwidth]
{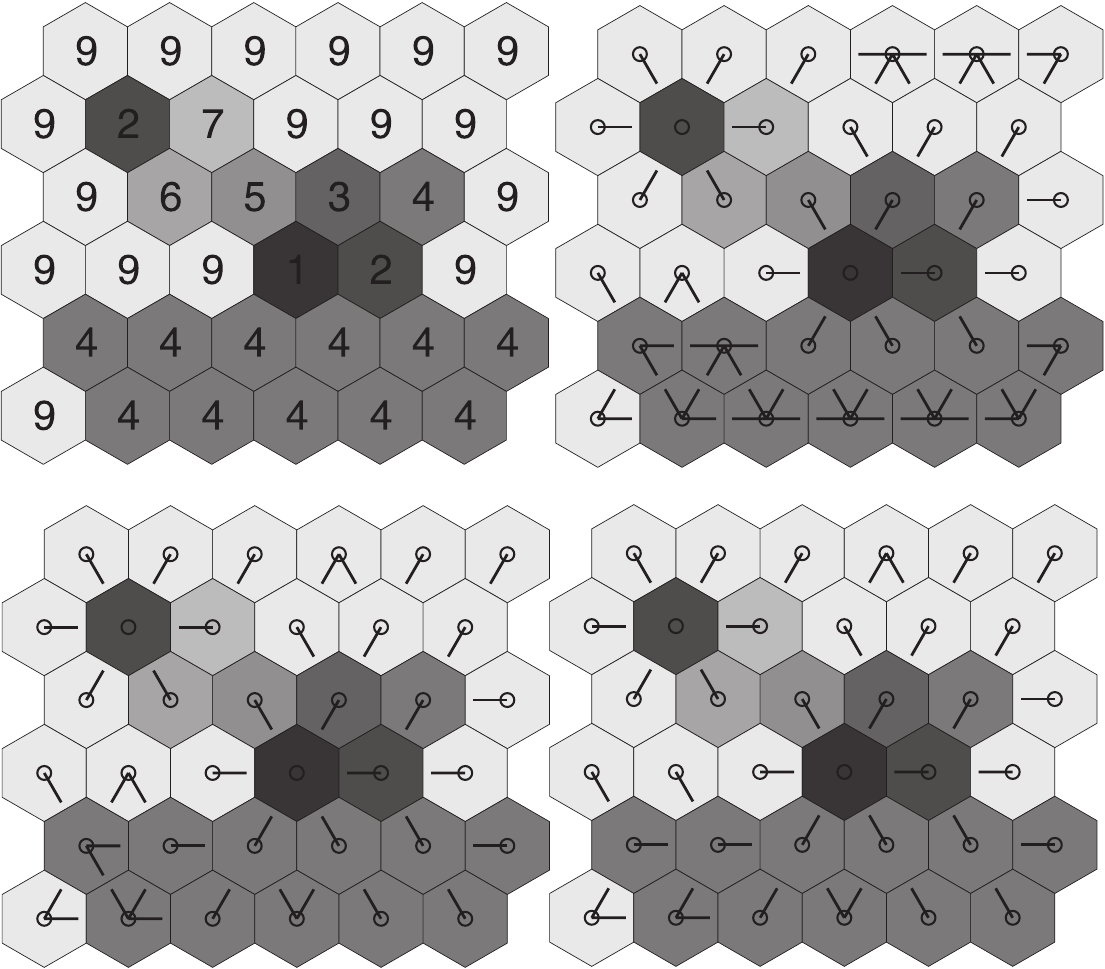}%
\caption{Successive prunings of the arrows.\ }%
\label{wfal90}%
\end{center}
\end{figure}

\subsection{Complexity}

After the initial detection and labeling of the minima, the algorithm needs a
number of iterations equal to the largest distance between a pixel in a
catchment basin and its regional minimum.\ Each iteration consists in the
combination of the adaptive erosion and dilation.\

\subsection{The problem of the plateaus}

\begin{figure}
[ptb]
\begin{center}
\includegraphics[width=0.66\textwidth]
{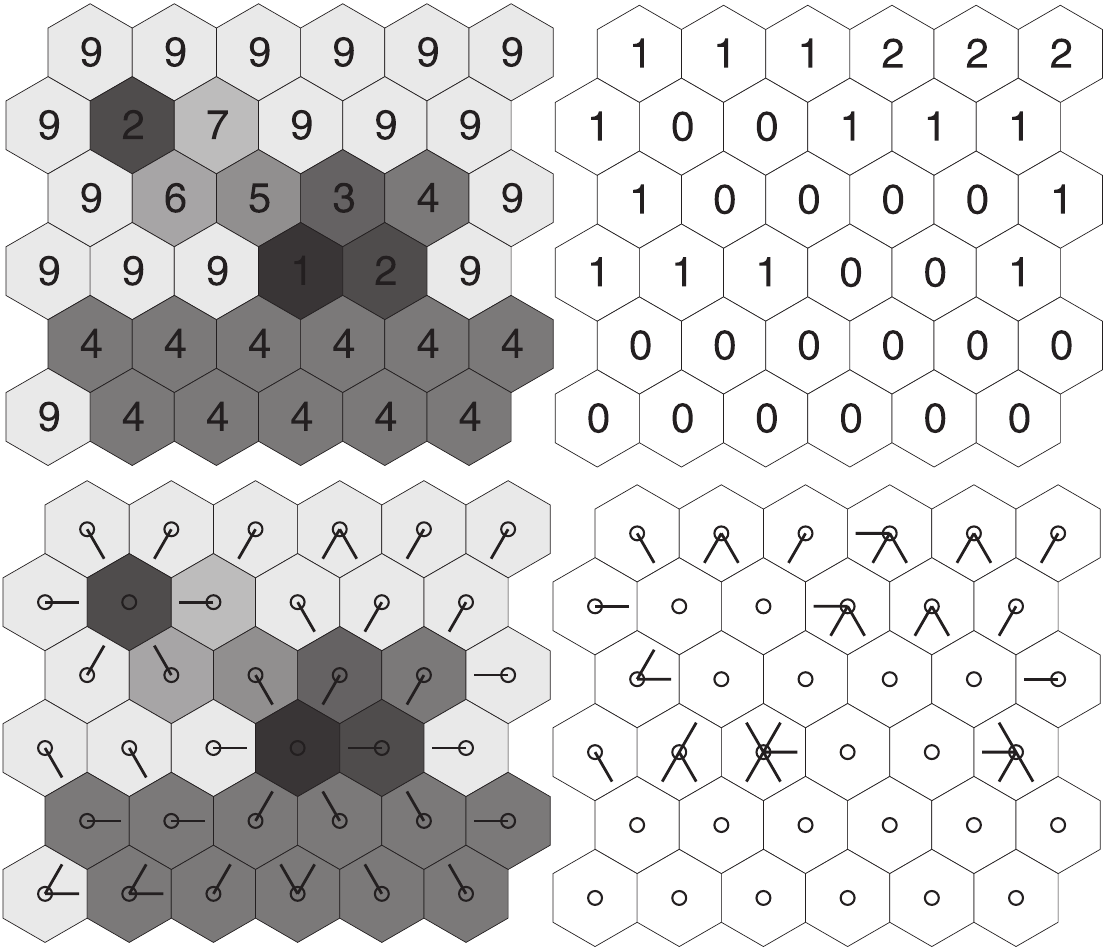}%
\caption{Top: On the left a topographic surface and on the right the geodesic
distance to its lower boundary on each plateau. \newline Bottom: On the left
the arrows of the steepest drainage graph and on the right the arrows
associated to the distance function.\ }%
\label{wfal91}%
\end{center}
\end{figure}

The plateaus pose a particular problem to all watershed algorithms which only
consider local neighborhoods.\ As a matter of fact, a drop of water falling
on a plateau has no clear direction for reaching the nearest regional
minimum.\ The classical solution consists in constructing a geodesic distance
transform to the lowest neighbors of the plateau and to follow the steepest
descent line on this function. Fig. \ref{wfal91}.1.1 shows a topographic
surface containing several plateaus with value $9.$ The geodesic distance
within each plateau to its lower boundary is illustrated in fig.\ \ref{wfal91}.1.2.\ 
This produces a topographic surface on which we may compute the drainage
graph, as illustrated in fig.\ \ref{wfal91}.2.2.\ By comparison, the steepest
drainage graph produced by our algorithm is more selective and has less arrows
It\ is illustrated in fig. \ref{wfal91}.2.1. Furthermore no special treatment is
required for dealing with the plateaus : they are treated as any other part of
the topographic surface.\ 

\begin{remark}
Some watershed algorithms use arrows for the construction of the watershed.
F.Maisonneuve is the first to assign arrows to all pixels with lower neighbors, and
then iteratively completing the arrowing inside plateaus \cite{FrancisM}.\ F.\ Lemonnier, in a hardware implementation of the
watershed, constructs separately the arrows of the drainage graph and those
of the geodesic distance within the plateaus, before regrouping both and
completing the watershed construction \cite{lemonth}.
\end{remark}

\section{Illustration on a real image}

\begin{figure}
[ptb]
\begin{center}
\includegraphics[width=0.99\textwidth]
{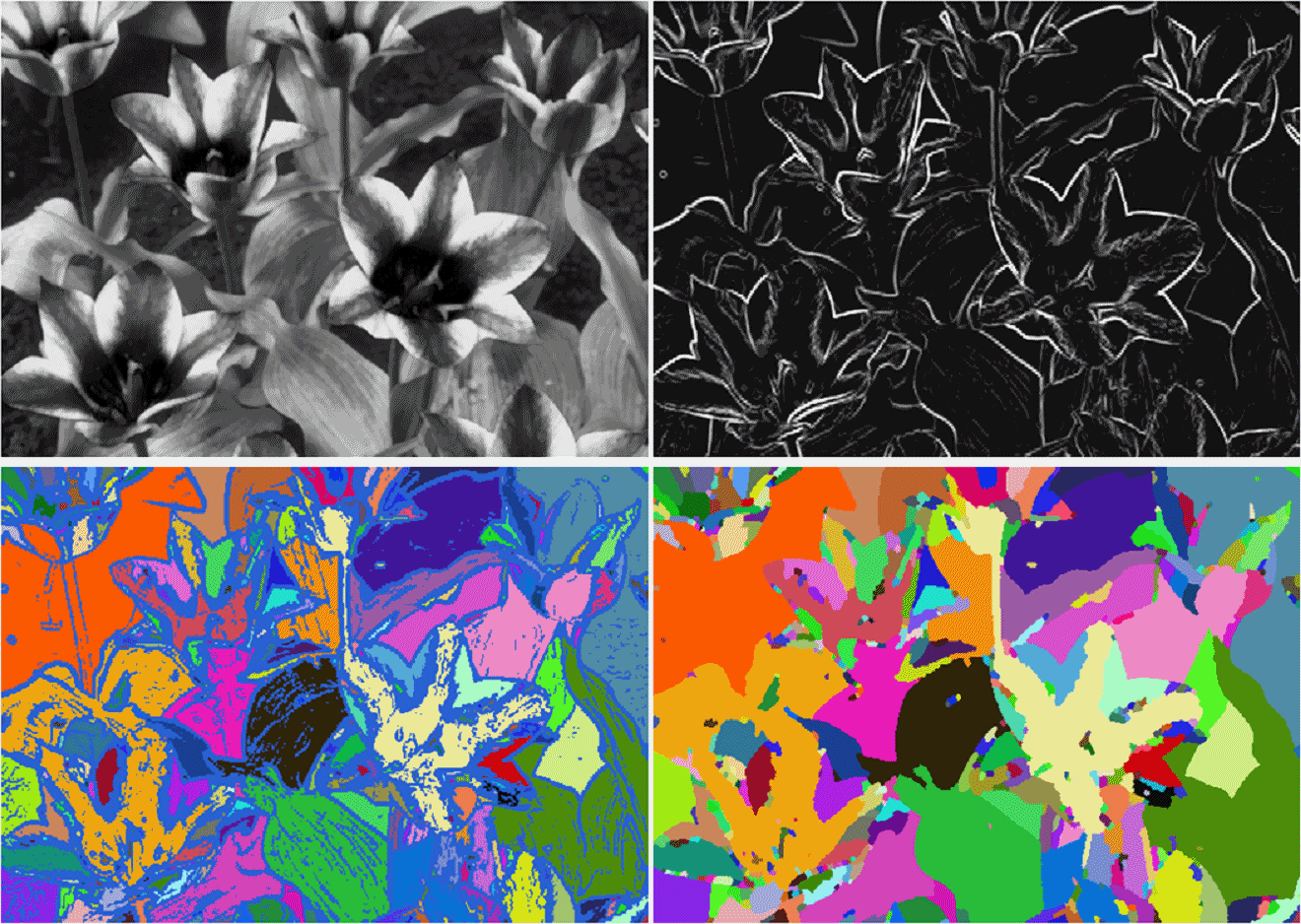}%
\caption{Initial image, gradient, labeled regional minima and labeled
catchment basins.\ }%
\label{wwzpe1}%
\end{center}
\end{figure}

\begin{figure}
[ptb]
\begin{center}
\includegraphics[width=0.99\textwidth]
{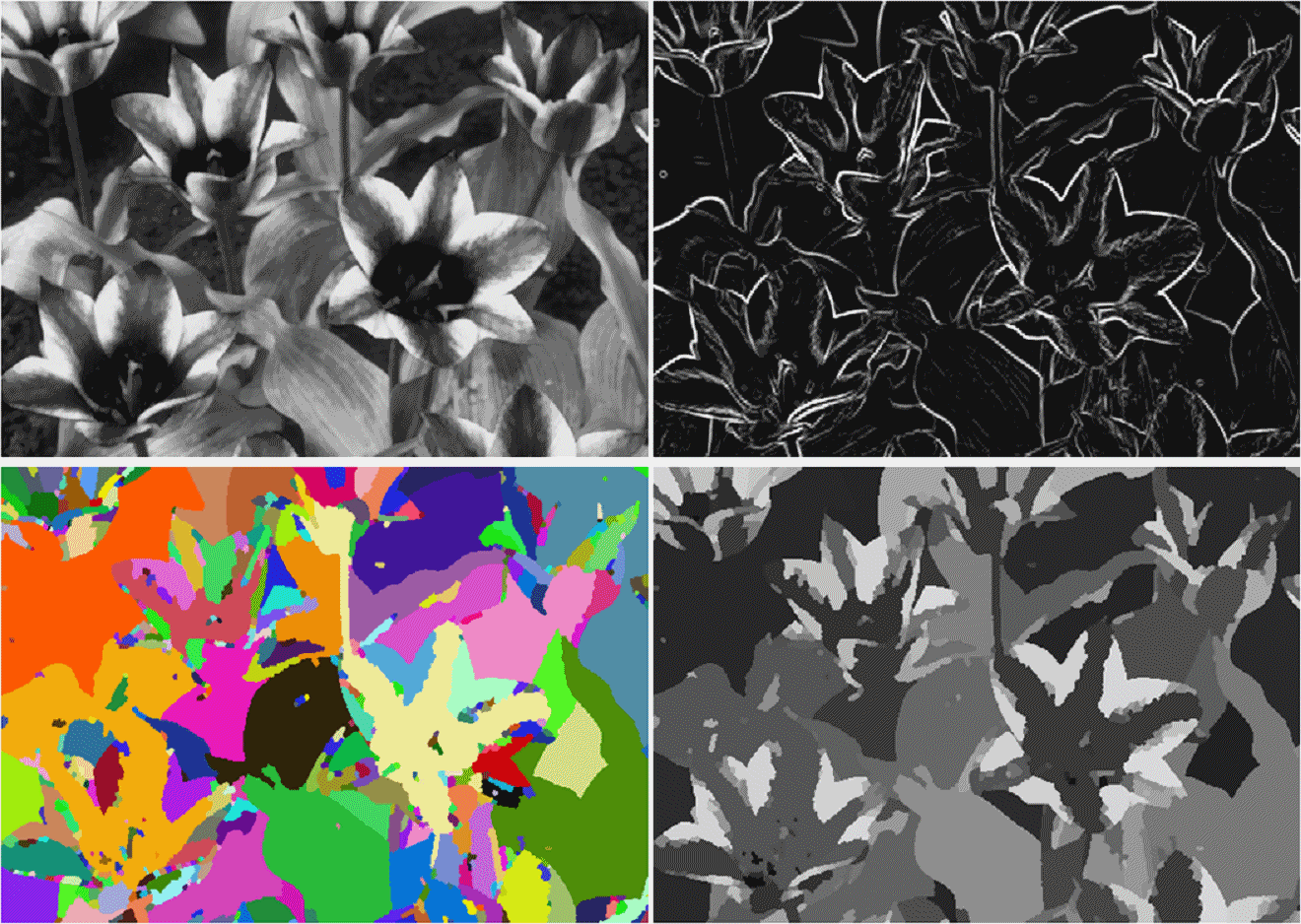}%
\caption{Initial image, gradient, labeled regional minima and as final result
the catchment basins containing each the mean gray tone of the initial
image.\ }%
\label{wwzpe2}%
\end{center}
\end{figure}

\begin{figure}
[ptb]
\begin{center}
\includegraphics[width=0.75\textwidth]
{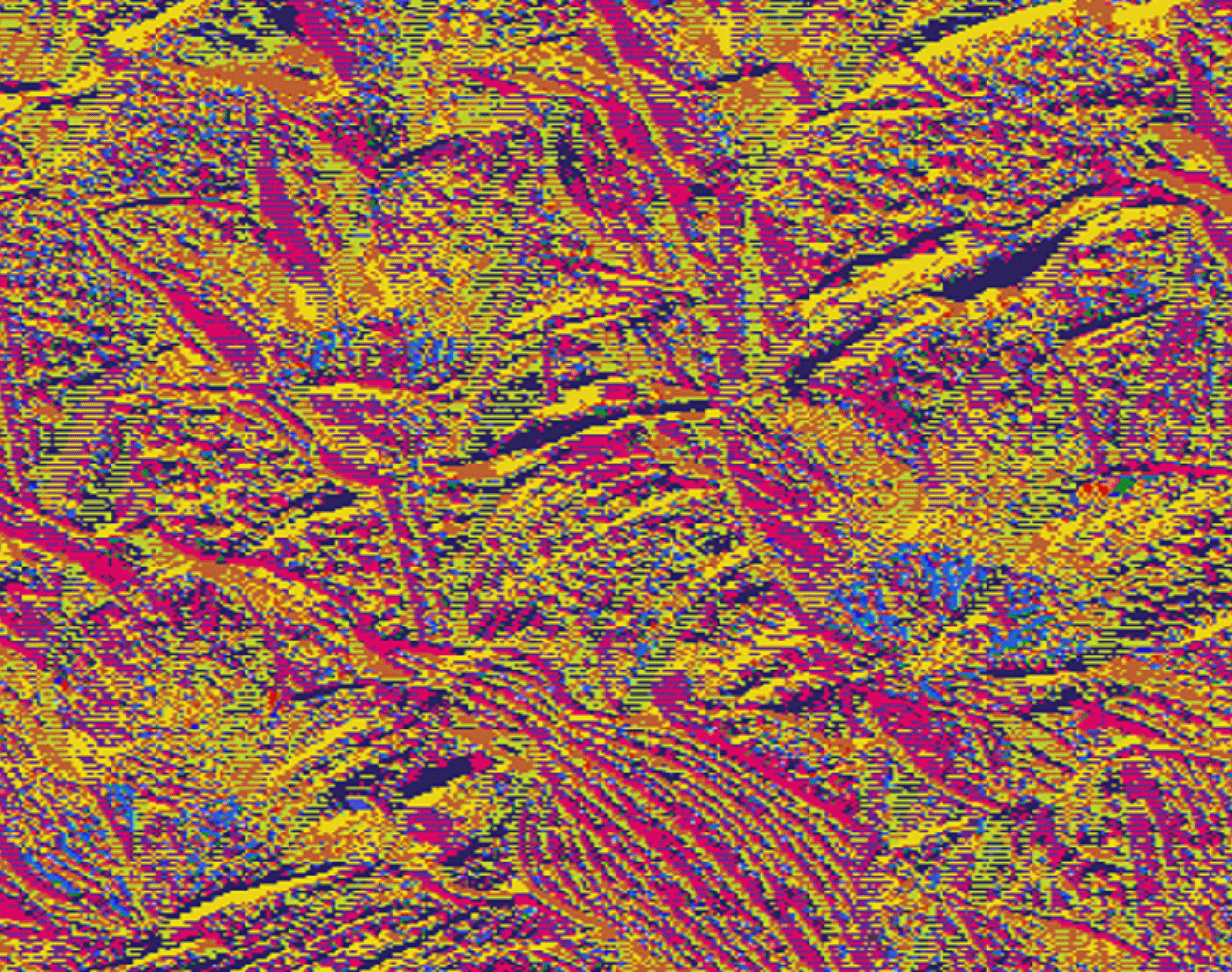}%
\caption{Central part of the arrow image of the steepest flooding graph represented in false
color.\ }%
\label{fleches}%
\end{center}
\end{figure}

The figures \ref{wwzpe1} and \ref{wwzpe2} illustrate the method on a real
image. They contain in the top row a gray tone image and its gradient.\ The
bottom row contains on the left the labeled regional minima and on the right
the associated catchment basin.\ In fig. \ref{wwzpe1}.2.2  they hold the same
labels as the minima they contain.\ Fig. \ref{wwzpe2}.2.2 is a mosaic image where each catchment
basin takes the mean gray tone of the initial image in this region.\ Fig.
\ref{fleches} represents the final drainage graph and shows the image of the
arrows encoded in false color.

%

%

\subsection{Contrast with the flooding algorithms : the watershed of a digital
elevation map}

\begin{figure}
[ptb]
\begin{center}
\includegraphics[width=0.9\textwidth]
{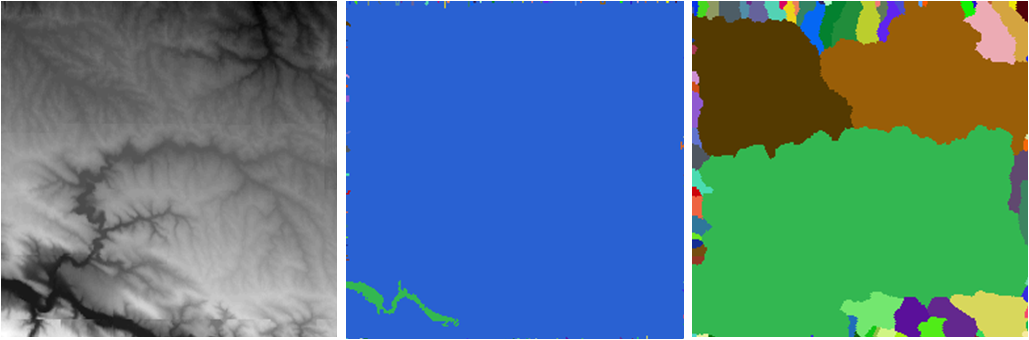}%
\caption{The gray tone image represents a digital elevation map.\ The central
image shows all regional minima touching the boundary of the image.\ The right
image presents the catchment basins.}%
\label{reliefsw}%
\end{center}
\end{figure}

Fig. \ref{reliefsw}.1.1  presents a digital elevation map of an existing
landscape : each gray tone represents an altitude.\ Due to sensing errors,
there are spurious regional minima in the topographic surface.\ As the
rivers leave the image in the direction of the see, the only minima which make
sense are those on the boundary of the image.\ A marker image is produced,
equal to the relief on the boundary of the image and to $\infty$
elsewhere.\ The highest flooding of the relief under this mask has all its
minima touching the boundary (see fig. \ref{reliefsw}.1.2 ); all other spurious
minima due to sensing errors have been suppressed. lexicographic watershed
produces the partition in catchment basins illustrated by fig.\ \ref{reliefsw}.1.3.\ 
Its catchments are the real catchment basins of the topographic surface,
containing each a river, appearing as a thalweg line.\ %

As announced earlier, the labels are propagated along lines of steepest
descent ; at each iteration they progress one step further. When they reach
the top of a drainage path, they stop.\ This may be clearly seen in fig.
\ref{relief1} showing the extension of the catchment basins after 20, 40, 60,
80, 100\ and 120 iterations of the elementary step of the algorithm
(adaptive erosion of gray tones, dilation of labels and pruning of
arrows).\ The construction of the catchment basins touching the lower boundary
is achieved after a low number of iterations as they are small.\ They stop
growing, independently of the adjacent catchment basins.\ The adjacent
catchment basin, associated to the largest river reaches them after many more iterations.\ 

This is in absolute contrast with the flooding algorithms which construct the
catchment basins as attraction zones of the regional minima.\ For such
algorithms, a catchment basin stops growing only when it arrives in contact
with another catchment basin. The most well known such algorithms implement
the topographic distance, as recalled below.

\subsubsection{The flooding algorithms}

Steepest descent lines are followed both by a drop of water gliding downwards
or by an increasing flood, which invades a topographic surface. The great
majority of watershed algorithms mimic the flooding of a topographic surface.
The regional minima are the sources, the level of the flood is uniform : as it
increases, the new lakes are created and old ones expand. The algorithm cares
that two lakes containing distinct minima do not merge.\ As a matter of fact,
the floodings progresses according lines of steepest descent of the surface
$;$ furthermore, these lines can be modeled as the geodesics of the topographic distance
function \cite{Najman199499},\cite{meyer94}.\ The catchment basins represent
the Vorono\"{\i} tessellation of the regional minima for the topographic
distance to the minima, whose altitude has been set to 0.\ The following
classical algorithm is derived from this definition.\ It uses a hierarchical
queue which correctly organizes the flooding in the presence of plateaus
\cite{meyer91}: a hierarchical queue is a series of first in first out queues,
with a priority order between the queues.\ The pixels are put in the queue
corresponding to their altitude ; pixels with a lower altitude having a higher
priority than pixels with a higher altitude.\ Furthermore, pixels of the same
queue are treated on a first in first out basis : like that the pixels of
plateaus of uniform altitude are treated in the order of increasing distances
to the boundary of the plateaus. The algorithm constructing the zones of
influence of the minima for the topographic distance is the following:

Label the nodes of the minima and introduce them in a hierarchical queue HQ each with a priority
equal to their weight.\ 

As long as the HQ\ is not empty, extract the node $j$ with the highest
priority from the HQ:

For each unlabeled neighboring (on the flooding graph) node $i$ of $j:$

\qquad* $label(i)=label(j)$

\qquad* put $i$ in the queue with priority $\ \nu_{i}$%

\begin{remark}
The hierarchical queue structure does most of the job for a proper
flooding.\ Lower pixels are flooded before higher ones, due to the hierarchy
between the queues ; pixels near the boundary of a plateau are treated before
more inwards pixels, due to the first in first out administration of each
queue. This management of flooding saves the day for the short-sighted
watershed algorithms, which accept too many downstream trajectories, resulting
in thick watershed zones. The hierarchical queue structure (or other similar
structures) helps making a proper division of these thick watershed zones. The
formulation of the algorithm is deceptive, as it gives an elegant mathematical
definition of catchment basins, but the fact that they largely overlap is
hidden ; the illustrations look convincing, however, the credit for this
quality is to be given more to the data structure for implementing them as to
the way they are defined.\ The algorithms with the highest myopia are those based on
the flooding ultrametric distance \cite{waterfallsbs},\cite{4564470}.
\end{remark}%

\begin{figure}
[ptb]
\begin{center}
\includegraphics[width=0.99\textwidth]
{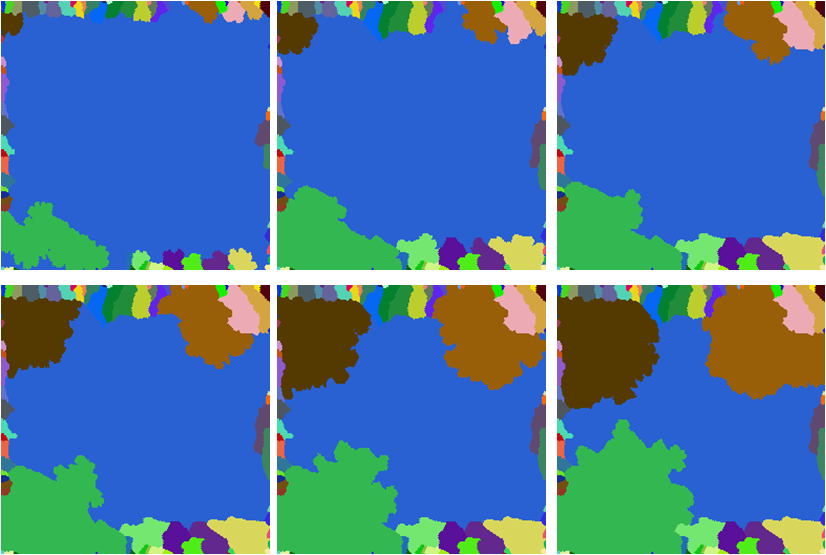}%
\caption{Extension of the catchment basins after 20, 40, 60, 80, 100\ and 120
iterations of the combined adaptive erosion of gray tones, pruning and
dilation of labels.}%
\label{relief1}%
\end{center}
\end{figure}
%

\begin{figure}
[ptb]
\begin{center}
\includegraphics[width=0.99\textwidth]
{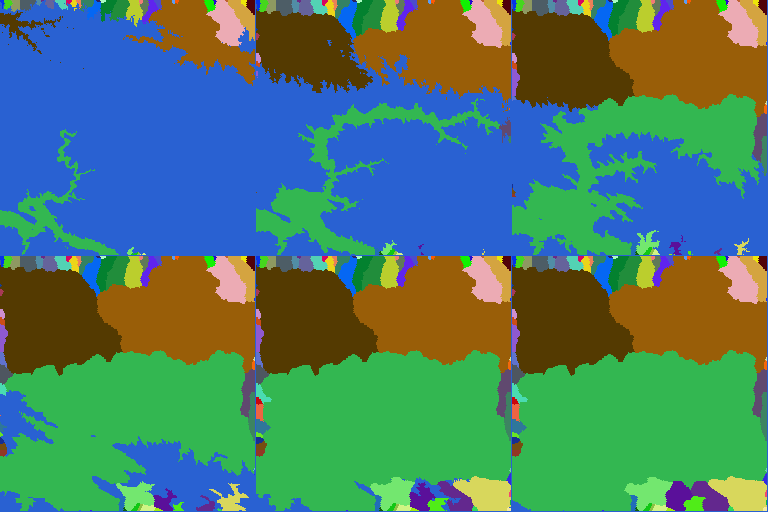}%
\caption{Construction of the catchment basins with an algorithm based on
uniform flooding}%
\label{wzpe4}%
\end{center}
\end{figure}

The grows of catchment basins for both algorithms may now be compared.\ Fig.
\ref{relief1} shows the extension of the catchment basins after 20, 40, 60,
80, 100\ and 120 iterations of the combined adaptive erosion of gray tones,
pruning and dilation of labels. One remarks a striking difference with the
classical algorithm for constructing the watershed which is based on the
simulation of a flooding, where a flood starting from the regional minima
grows with a uniform altitude and progressively invades the topographic
surface. Fig. \ref{wzpe4} precisely shows the progressive construction of the
catchment basin based on the flooding distance; the unflooded part appears in
dark blue, the flooded parts appears as colored labels.\ The successive
levels of flooding represented are 75, 105, 135,\ 170, 205 and 240 (gray-tones
on a scale $[0,255]$). The flooding algorithm is a greedy shortest distance
algorithm based on the topographic distance.\ Each catchment basin stops
growing everywhere it meets another catchment basin. With the steepest path
algorithm on the contrary, the catchment basins are dilated at each iteration
by a dilation of size one and they stop growing when they have reached their
full extension, have they reached another catchment basin or not.\ As a
consequence, if we suppress the label of a minimum at initialization, the
catchment basin of this minimum will remain empty.\

\section{Downstream trajectories of a drop of water}

\subsection{The algorithm for downstream propagation}

\begin{figure}
[ptb]
\begin{center}
\includegraphics[width=0.6\textwidth]
{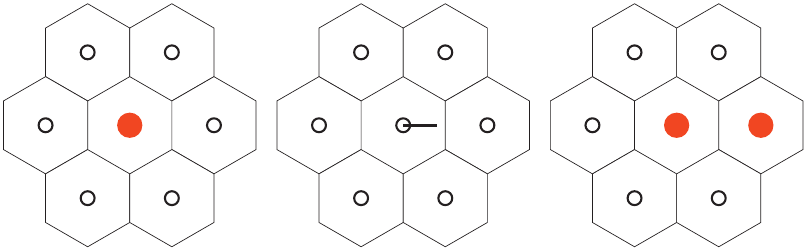}%
\caption{Downstream following of a drop of water: a labeled node is expanded
in the directions of its arrows to its neighboring nodes.}%
\label{wfal100}%
\end{center}
\end{figure}

The successive prunings of a drainage graph leaves a minimal graph containing
the steepest downstream trajectories for each node. Given a few labeled
starting points, it is possible to follow the downstream trajectories, simply
by following the downstream arrows. Fig. \ref{wfal100} illustrates the
operator which is used. If a pixel $p$ has a label (fig. \ref{wfal100}.1) and
an arrow towards a neighboring pixel $q,$ (fig. \ref{wfal100}.2) then  the
label is propagated from  $p$ to $q$ (fig. \ref{wfal100}.3). If $q$ has
already a label, the maximum of both labels is chosen.\ This expansion is
repeated until stability.

\begin{figure}
[ptb]
\begin{center}
\includegraphics[width=0.66\textwidth]
{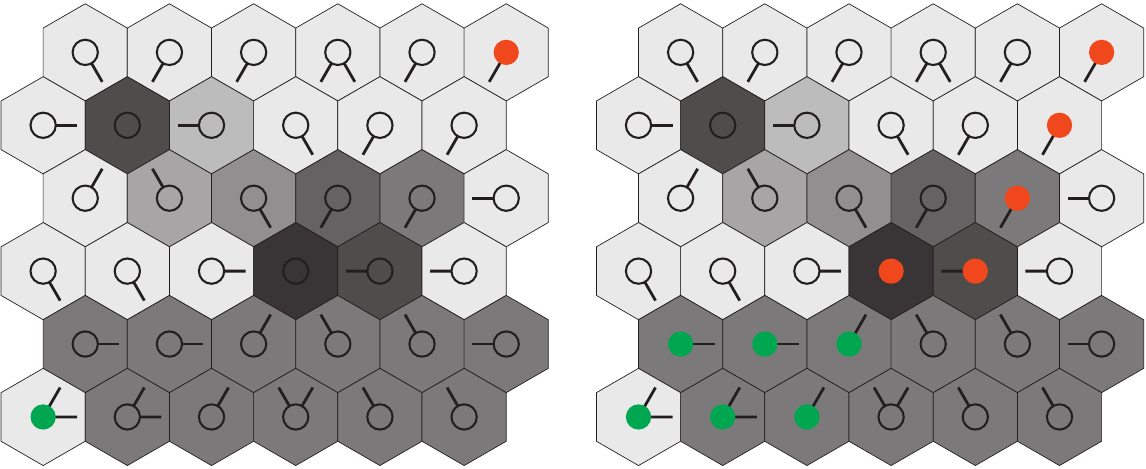}%
\caption{Two starting nodes are chosen and assigned a label.\ On the right
figure, the downwards trajectories are illustrated.\ }%
\label{wfal99}%
\end{center}
\end{figure}

In fig. \ref{wfal99}.1 two starting points are chosen, and assigned distinct labels, represented
respectively by a red and a green dot. After downwards propagation following the arrows,  both
trajectories appear as labeled nodes in fig. \ref{wfal99}.2, each of them reaching a regional minimum.\ %

\begin{figure}
[ptb]
\begin{center}
\includegraphics[width=0.99\textwidth]
{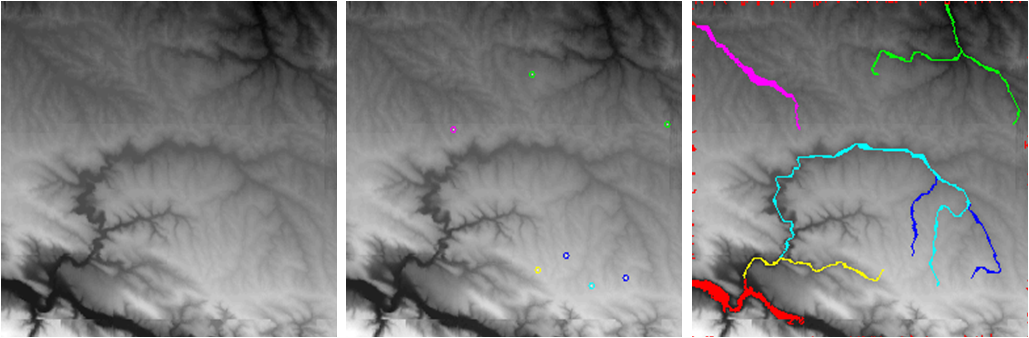}%
\caption{After construction of the arrows of the DEM, a number of starting
points are chosen and labeled (central image) and the downstream trajectories
of a drop of water falling on these points highlighted in the right image.}%
\label{rivers}%
\end{center}
\end{figure}

This method is applied on the same DEM\ image in fig. \ref{rivers}.1 after the
construction of the steepest drainage graph. A number of positions are chosen
by hand in  in fig. \ref{rivers}.2, where a drop of water will start its downwards
trajectory, highlighted in the image on the right. The drop of water duly
glides downwards until it meets and follows a river and ultimately reaches the
boundary of the image in the direction of the sea, as illustrated in fig. \ref{rivers}.3. Each river keeps the label
of its highest source and keeps this label unless it meets another label which
is higher.\ The connected components colored in red and touching the boundary of the image
represent the regional minima of the image.%

\subsection{Application to the detection of fibers, cracks, thalweg lines.}

\begin{figure}
[ptb]
\begin{center}
\includegraphics[width=0.99\textwidth]
{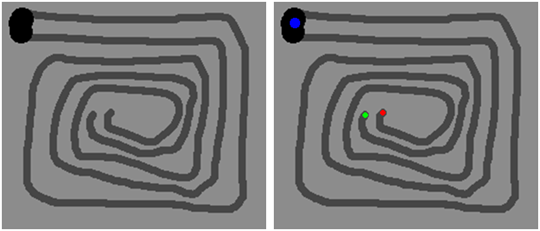}%
\caption{Left : Two spirals on a bright background ending in a dark
minimum\newline Right : A red and and a green dot mark the extremities of the
spiral. A blue dot the minimum.\ }%
\label{spiral6}%
\end{center}
\end{figure}

\begin{figure}
[ptb]
\begin{center}
\includegraphics[width=0.99\textwidth]
{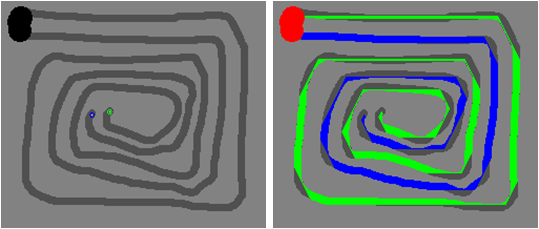}%
\caption{The spiral stripes  have a constant altitude, that is they are
plateaus, without internal structure for centering the trajectories, which
appear at some places as large as the stripes.\ }%
\label{spiral4}%
\end{center}
\end{figure}

Fig. \ref{spiral6} contains two spirals, intricated one in another, ending
with a dark regional minimum.\ The background is brighter than the spiral. A
red dot and a green dot have been put on the other extremity of both
spirals.\ The regional minimum has been marked by a blue dot.\ The spirals
are plateaus with a uniform gray-tone, with one lower boundary in the
regional minimum.\ The steepest descent trajectories taking their origin in
the red and green tots are represented in fig.\ref{spiral4}.\ They start from
the red and green dots and follow the stripes until they reach the
minimum.

\begin{figure}
[ptb]
\begin{center}
\includegraphics[width=0.9\textwidth]
{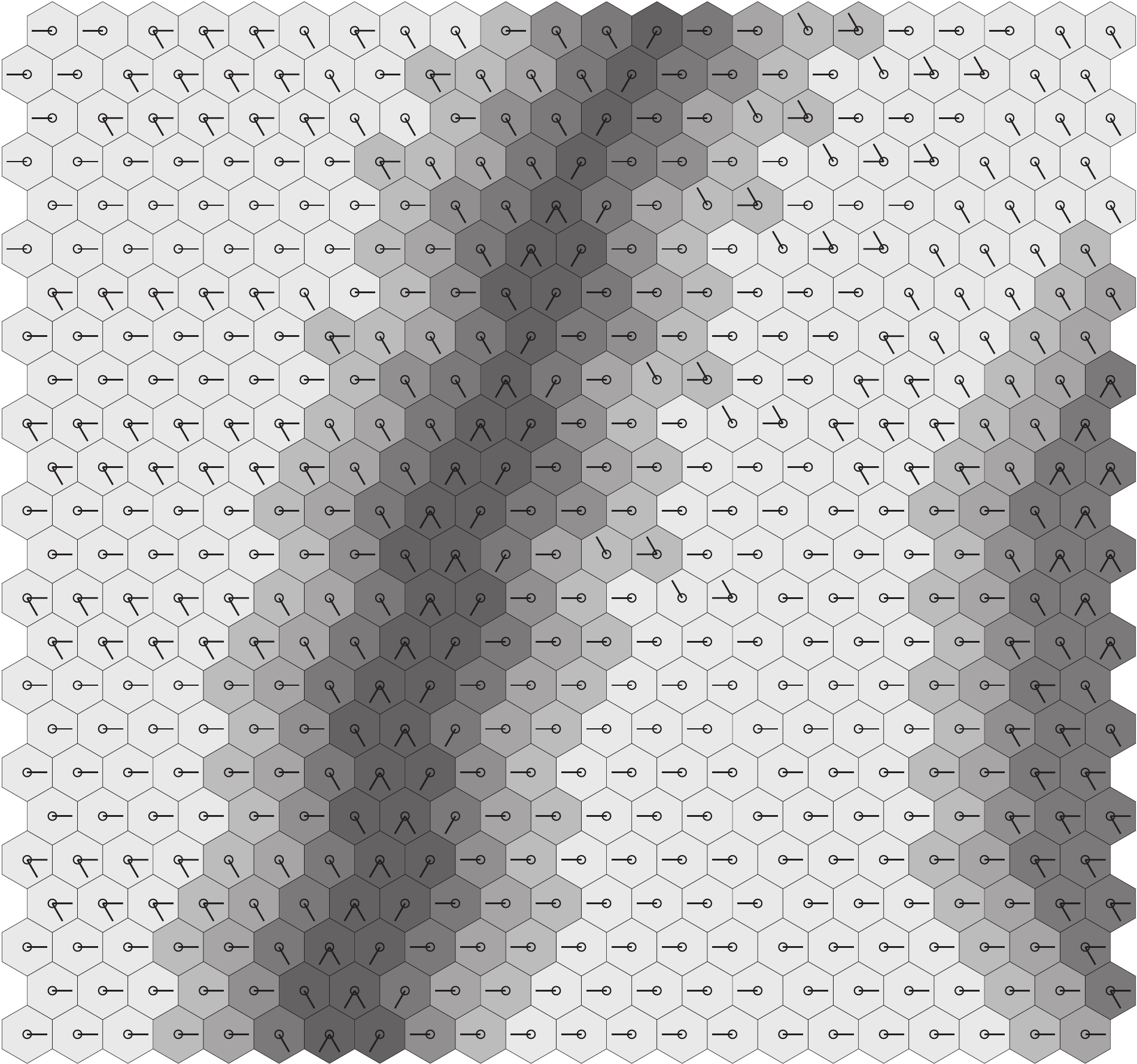}%
\caption{The arrows of the steepest drainage graph after constructing the
geodesic distance of each node of a plateau to the lower border of the
plateau.\ }%
\label{spiral0}%
\end{center}
\end{figure}

\begin{figure}
[ptb]
\begin{center}
\includegraphics[width=0.99\textwidth]
{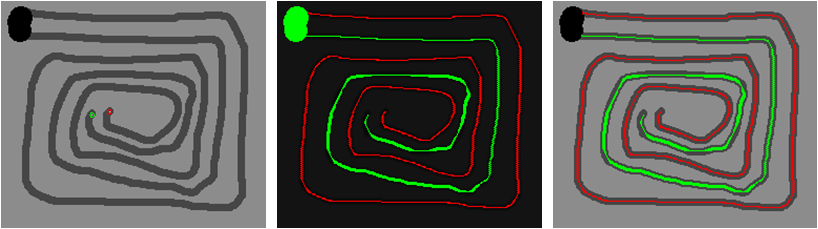}%
\caption{Replacing each strip by a distance transform to its lowest boundary
permits to center the downwards trajectories inside the stripes.}%
\label{spiral5}%
\end{center}
\end{figure}

The spiral stripes in fig.\ \ref{spiral4}  are plateaus of constant altitude,
without internal structure for centering the trajectories, which appear at
some places as large as the stripes in the right image. For a better centering
of the trajectories, we have transformed the binary image into a gray tone
image by constructing a geodesic distance to the lower borders of the
plateaus.\ The arrows of the steepest drainage graph within a small square crossing a stripe are shown in
fig.\ \ref{spiral0}.\ The stripes are well centered and the trajectory along
their thalweg well delineated.\ On this new relief, the detected trajectories
are much thinner (fig.\ref{spiral5})%

\subsubsection{Comparison with shortest path algorithms}

The classical algorithms for following and highlighting thin and elongated
dark structures (let us call it fiber) in a noisy bright background rely on
shortest path algorithms (for instance cracks in a porous medium, hairs, glass
fibers, vessels in 2 or 3 dimensions etc.). As the gray tone along the fiber
is darker than in the background, the integral of gray tone along the fiber is
smaller than along a path with the same length lying in the brighter
background. This integral may be seen as a weighted distance transform. The
method consists in computing the weighed distance along the fiber, first in
one direction, then in the opposite direction ; the sum of both distance
transforms is then minimal along the fiber.\ The method has been first
proposed for detecting cracks in porous material \cite{VincentJeulin}%
,\cite{Vincentminpaths}.\ The method works well if the cracks or fibers or
more or less rectilinear.\ 

However, if the fiber is tortuous, like the spirals above, then a shortest
path algorithm between the regional minimum and the  red dot (resp. green) 
will find shortcuts and will not follow the spiral. The classical
solution consists in a stepwise progression : one progresses along the fiber
on a short distance, so as to remain on it and initiates a new progression at
the arrival point.\ Like that, little jump after little jump, one progresses.
The length of the jump depends upon the contrast between fiber and background
\cite{CohenDeschamps}.\ 

On the contrary, using the drainage graph does nor present this weakness : the
trajectories are delineated and based on a long range lexicographic distance
which makes it possible to follow them completely from top to down,
until a regional minimum is met.\ Contrarily to the shortest path algorithm,
the algorithm has not to be used stepwise, does not depend upon the contrast
between foreground and background. It is also able to follow several fibers at
the same time.\ However, it is certainly more sensitive to noise or missing
data than the shortest path algorithms.\

%

%

\section{Discussion}

The present work is the first implementation and validation of a more
ambitious work (under publication) where we consider node or edge weighted
graphs and show that they are in fact equivalent from the point of view of the
watershed construction. We show how to complete the missing weights, such that the
weight of a node is the smallest weight of its adjacent edges, i.e. an erosion from
edges to nodes, and the weight of an edge the highest weight of its extremities,
i.e. a dilation from nodes to edges.\ Erosion and dilation being adjunct from
each other, the associated opening $\gamma_{e}$ is particularly interesting :
the edges invariant by this opening are all edges which are the lowest edge of
one of their extremities.\ This indicates that a flood coming from this higher
extremity may pass through this edge.\ 

We next define two erosions, assigning to a node weight(resp.\ edge) the minimal
weight of its adjacent nodes (resp.\ edges).\ After eroding edges and nodes, a
new graph is created which is not necessarily invariant by the opening
$\gamma_{e}.\ $We prune it by suppressing all edges which are not invariant
by $\gamma_{e}.$ Repeating this operation, erosion/pruning until stability,
produces a minimal graph, where only the steepest paths remain.

In the present work also we use successive erosions of the drainage graph and
prunings so as to get a drainage graph containing only the steepest paths. The
difference lies in the implementation of the algorithm. Here we have no
possibility to assign weights to edges and we have to use oriented arcs for
representing the drainage graph. As the neighborhood structure is fixed, the
repartition of arrows having a pixel as extremity may be encoded as a binary
number of length equal to the number of neighbors.\ 

The present algorithm could also be applied to an edge weighted graph after transforming it into a node weighted graph, obtained by assigning to
each node a weight equal to the smallest weight of its adjacent edges.

\section{Conclusion}

We have presented a particularly simple parallel algorithm for the
construction of the watershed of gray tone images. As it is defined for node
weighted graphs, it may be adapted for any type of grid and any number of
dimensions.\ It extracts from a node weighted graph the smallest drainage
graph possible, without any arbitrary choices. Arbitrary choices only remain in the propagation of labels, 
in the case where a given node is linked with two distinct minima by 2 drainage paths with
exactly the same weights in the same order.

The algorithm is based on simple parallel neighborhood operators, which are
particularly suitable for an implementation in special hardware or in graphics processors.\ 

Many variants of this algorithm can be thought of:

\begin{itemize}
\item It may be optimized in different ways.\ For instance as soon a pixel has
only one arrow, its label can arrive only from this direction. In terms of
graphs, the corresponding edge could be contracted. In terms of images, one
could connect such pixels with a "union-find" type of algorithm.

\item The algorithm could also be used as a pre-processing step, with only a
small number of iterations performed, in order to significantly prune the
graph. The arrows may then be inverted (if $i$ arrows $j$ in an image$,$ $j$
arrows $i$ after inversion of the arrows). Like that any classical algorithm
based on hierarchical queues would consider propagating the labels only in the
direction of the inverted arrows.

\item It may also be used as a preprocessing step in some applications where a
high speed is required : apply a fixed number of steps of pruning, and then
suppress all possibilities of choice, by keeping only one arrow for each node.
The algorithm concludes by extracting and labeling the connected components. 
(A.Bieniek, in \cite{Bieniek2000907}, keeps an arrow only between each node and one of its lowest
neighbors and then labels the resulting graph.) 
\end{itemize}

The lexicographic watershed selects a minimal set of downstream
directions.\ Furthermore, the selectivity of the pruning allows to construct
long range downstream trajectories, even in the presence of high tortuosity.\

---------------------------------------------------
\bibliographystyle{plain}
\bibliography{C:/swp55/TCITeX/BibTeX/bib/biblio2,C:/swp55/TCITeX/BibTeX/bib/wshed,}

\newcommand{\noopsort}[1]{} \newcommand{\printfirst}[2]{#1}
  \newcommand{\singleletter}[1]{#1} \newcommand{\switchargs}[2]{#2#1}
\begin{thebibliography}{10}

\bibitem{berge85}
C.~Berge.
\newblock {\em Graphs}.
\newblock Amsterdam: North Holland, 1985.

\bibitem{beucher79}
S.~Beucher and C.~Lantu{\'e}joul.
\newblock Use of watersheds in contour detection.
\newblock In {\em Proc. Int. Workshop Image Processing, Real-Time Edge and
  Motion Detection/Estimation}, 1979.

\bibitem{Bieniek2000907}
Moga~A. Bieniek, A.
\newblock An efficient watershed algorithm based on connected components.
\newblock {\em Pattern Recognition}, 33(6):907--916, 2000.

\bibitem{4564470}
J.~Cousty, G.~Bertrand, L.~Najman, and M.~Couprie.
\newblock Watershed cuts: Minimum spanning forests and the drop of water
  principle.
\newblock {\em Pattern Analysis and Machine Intelligence, IEEE Transactions
  on}, 31(8):1362 --1374, aug. 2009.

\bibitem{fusiongraphs}
Jean Cousty, Michel Couprie, Laurent Najman, and Gilles Bertrand.
\newblock Grayscale watersheds on perfect fusion graphs.
\newblock In Ralf Reulke, Ulrich Eckardt, Boris Flach, Uwe Knauer, and Konrad
  Polthier, editors, {\em Combinatorial Image Analysis}, volume 4040 of {\em
  Lecture Notes in Computer Science}, pages 60--73. Springer Berlin /
  Heidelberg, 2006.

\bibitem{CohenDeschamps}
Thomas Deschamps and Laurent~D. Cohen.
\newblock Fast extraction of minimal paths in 3d images and applications to
  virtual endoscopy.
\newblock {\em Medical Image Analysis}, 5(4):281 -- 299, 2001.
\newblock <ce:title>Soft Tissue Deformation</ce:title>.

\bibitem{IftLotufo}
A.X. Falcao, J.~Stolfi, and R.~de~Alencar~Lotufo.
\newblock The image foresting transform: theory, algorithms, and applications.
\newblock {\em Pattern Analysis and Machine Intelligence, IEEE Transactions
  on}, 26(1):19 -- 29, jan 2004.

\bibitem{gondranminoux}
M.~Gondran and M.~Minoux.
\newblock {\em Graphes et Algorithmes}.
\newblock Eyrolles, 1995.

\bibitem{grimaud92}
M.~Grimaud.
\newblock New measure of contrast~: dynamics.
\newblock {\em Image Algebra and Morphological Processing III, San Diego CA,
  Proc. SPIE}, 1992.

\bibitem{lemonth}
F.~Lemonnier.
\newblock {\em Architecture Electronique D\'edi\'ee aux Algorithmes Rapides de
  Segmentation Bas\'es sur la Morphologie Math\'ematique}.
\newblock PhD thesis, E.N.S. des Mines de Paris, 1996.

\bibitem{FrancisM}
F.~Maisonneuve.
\newblock Sur le partage des eaux.
\newblock {\em Technical Report-Centre of Mathematical
  Morphology-Mines-ParisTech}, 1982.

\bibitem{waterfallsbs}
B.~Marcotegui and S.~Beucher.
\newblock Fast implementation of waterfalls based on graphs.
\newblock {\em ISMM05 : Mathematical Morphology and its applications to Signal
  Processing}, pages 177--186, 2005.

\bibitem{meyer91}
F.~Meyer.
\newblock Un algorithme optimal de ligne de partage des eaux.
\newblock In {\em Proceedings $8^{\underline{\grave{e}me}}$ Congr\`es AFCET,
  Lyon-Villeurbanne}, pages 847--857, 1991.

\bibitem{meyer92}
F.~Meyer.
\newblock Color image segmentation.
\newblock In {\em IEE Fourth International Conference on Image Processing and
  its Applications}, pages 303--306, April 1992.

\bibitem{meyer94}
F.~Meyer.
\newblock Topographic distance and watershed lines.
\newblock {\em Signal Processing}, pages 113--125, 1994.

\bibitem{floodingsegmey}
Fernand Meyer.
\newblock Flooding and segmentation.
\newblock In John Goutsias, Luc Vincent, and Dan~S. Bloomberg, editors, {\em
  Mathematical Morphology and its Applications to Image and Signal Processing},
  volume~18 of {\em Computational Imaging and Vision}, pages 189--198. 2002.

\bibitem{wshedhistory}
Fernand Meyer.
\newblock The watershed concept and its use in segmentation : a brief history.
\newblock {\em (arXiv:1202.0216v1)}, 2012.

\bibitem{Najman94}
L.~Najman.
\newblock {\em Morphologie Math\'ematique: de la segmentation d'images \`a
  l'analyse multivoque}.
\newblock PhD thesis, Universit\'e Paris-Dauphine, 1994.

\bibitem{saliency}
L.~Najman.
\newblock {Geodesic saliency of watershed edges and hierarchical segmentation.}
\newblock {\em IEEE Trans. Pattern Anal. Machine Intell}, 16(3):175--182, 1996.

\bibitem{Najman199499}
Laurent Najman and Michel Schmitt.
\newblock Watershed of a continuous function.
\newblock {\em Signal Processing}, 38(1):99 -- 112, 1994.
\newblock Mathematical Morphology and its Applications to Signal Processing.

\bibitem{Roerdink01thewatershed}
Jos B. T.~M. Roerdink and Arnold Meijster.
\newblock The watershed transform: Definitions, algorithms and parallelization
  strategies.
\newblock {\em Fundamenta Informaticae}, 41:187--228, 2001.

\bibitem{vachier95}
C.~Vachier.
\newblock {\em Extraction de Caract\'eristiques, Segmentation d'Image et
  Morphologie Math\'ematique}.
\newblock PhD thesis, E.N.S. des Mines de Paris, 1995.

\bibitem{vachier95b}
C.~Vachier and F.~Meyer.
\newblock Extinction values: A new measurement of persistence.
\newblock In I.~Pitas, editor, {\em 1995 IEEE Workshop on Nonlinera Signal and
  Image Processing}, pages 254--257, 1995.

\bibitem{VincentJeulin}
Jeulin~D. Vincent, L.
\newblock Minimal paths and crack propagation simulations.
\newblock {\em Acta Stereologica}, 8(2 II):487--494, 1989.

\bibitem{Vincentminpaths}
Luc Vincent.
\newblock Minimal path algorithms for the robust detection of linear features
  in gray images.
\newblock In {\em Proceedings of the fourth international symposium on
  Mathematical morphology and its applications to image and signal processing},
  ISMM '98, pages 331--338, Norwell, MA, USA, 1998. Kluwer Academic Publishers.

\bibitem{xisca2002}
F.~Zanoguera, B.~Marcotegui, and F.~Meyer.
\newblock A segmentation pyramid for the interactive segmentation of 3-d images
  and video sequences.
\newblock In John Goutsias, Luc Vincent, and Dan~S. Bloomberg, editors, {\em
  Mathematical Morphology and its Applications to Image and Signal Processing},
  volume~18 of {\em Computational Imaging and Vision}, pages 223--232. Springer
  US, 2002.

\end{thebibliography}

\end{document}